\documentclass[conference]{IEEEtran}
\IEEEoverridecommandlockouts
\usepackage{cite}
\usepackage{amsmath,amssymb,amsfonts}
\usepackage{algorithmic}
\usepackage{graphicx}
\usepackage{textcomp}
\usepackage{xcolor}

\usepackage{todonotes}
\usepackage[normalem]{ulem}

\usepackage[ruled,lined]{algorithm2e}
\usepackage{stmaryrd}
\usepackage{dsfont}
\usepackage{comment}
\usepackage[colorlinks]{hyperref}
\usepackage{amsthm}

\theoremstyle{plain}
\newtheorem{theorem}{Theorem}[section]

\newtheorem{conjecture}[theorem]{Conjecture}

\DeclareMathOperator{\Var}{Var}
\DeclareMathOperator{\argmax}{argmax}

\def\half{\frac{1}{2}}

\def\1{{\mathbf 1}}

\def\N{{\mathbb N}}

\def\R{{\mathbb R}}

\def\P{{\mathbb P}}
\def\E{{\mathbb E}}

\def\Fc{{\mathcal F}}

\def\Sc{{\mathcal S}}

\def\BibTeX{{\rm B\kern-.05em{\sc i\kern-.025em b}\kern-.08em
    T\kern-.1667em\lower.7ex\hbox{E}\kern-.125emX}}
    
\begin{document}

\title{ Sensitivity Analysis for Active Sampling, with Applications to the Simulation of Analog Circuits
\thanks{The authors R.C., F.G. and C.P acknowledge the support of the ANR-3IA ANITI (Artificial and Natural Intelligence Toulouse Institute).}
}

\author{\IEEEauthorblockN{CHHAIBI Reda}
\IEEEauthorblockA{\textit{Université Toulouse III - Paul Sabatier} \\
Toulouse, France \\
reda.chhaibi@math.univ-toulouse.fr \\
ORCID: 0000-0002-0085-0086}
\and
\IEEEauthorblockN{GAMBOA Fabrice}
\IEEEauthorblockA{\textit{Université Toulouse III - Paul Sabatier} \\
Toulouse, France \\
fabrice.gamboa@math.univ-toulouse.fr\\
ORCID: 0000-0001-9779-4393}
\and
\IEEEauthorblockN{OGER Christophe}
\IEEEauthorblockA{\textit{NXP Semiconductors} \\
Toulouse, France \\
christophe.oger@nxp.com}
\and
\IEEEauthorblockN{OLIVEIRA Vinicius}
\IEEEauthorblockA{\textit{NXP Semiconductors} \\
Toulouse, France \\
vinicius.alvesdeoliveira@nxp.com\\
ORCID: 0000-0003-2122-7733}
\and
\IEEEauthorblockN{PELLEGRINI Clément}
\IEEEauthorblockA{\textit{Université Toulouse III - Paul Sabatier} \\
Toulouse, France \\
clement.pellegrini@math.univ-toulouse.fr\\
ORCID: 0000-0001-8072-4284}
\and
\IEEEauthorblockN{REMOT Damien}
\IEEEauthorblockA{\textit{Université Toulouse III - Paul Sabatier}\\
Toulouse, France \\
damien.remot@math.univ-toulouse.fr}
\and
Authors are in alphabetical order.
}

\maketitle

\thispagestyle{plain}
\pagestyle{plain}


\tableofcontents

\medskip

\begin{abstract}
We propose an active sampling flow, with the use-case of simulating the impact of combined variations on analog circuits. In such a context, given the large number of parameters, it is difficult to fit a surrogate model and to efficiently explore the space of design features. By combining a drastic dimension reduction using sensitivity analysis and Bayesian surrogate modeling, we obtain a flexible active sampling flow. On synthetic and real datasets, this flow outperforms the usual Monte-Carlo sampling which often forms the foundation of design space exploration.
\end{abstract}


\section{\bf Introduction}

\subsection{\bf Context and literature}
\label{section:framework_and_contributions}

{ \bf Context:} Performances of integrated circuit are knowingly sensitive to a plurality of effects linked to the fabrication process or electrical stress via their impact on each device’s characteristics (process and mismatch variations, temperature, voltage, aging, etc). As such, today’s analog design flows are proposing various tools to simulate the impact of these effects. This has conducted the design flow to usually focus first on the impact of statistical variations of devices on the circuit performances using dedicated tools, and then once the circuit is tuned with respect to these, a next step can be to evaluate aging with other tools to verify that the impact of degraded devices stays within the remaining performance margins left after statistical simulation. If this sequential approach may be fine for circuits designed with mature nodes where aging’s impact on devices is normally orders of magnitude less compared to statistical variations, this becomes impractical in advanced nodes where it not rare to see the impact of aging more comparable to the one of statistical variations, as for instance mentioned in \cite{website}. In such cases, the only way would be to simulate the combined effect of these different causes, which leads to an explosion in number of simulations to run if for example at each statistical variations sample we execute the often multi-step stress conditions execution to add aging degradation. Methods helping to reduce the total number of simulations in such multi-variation domains scenarios are therefore highly desired to limit the exploding CPU cost they imply.

Indeed, Monte Carlo (MC) based analysis is necessary to provide a robust estimation assessment of the circuit's design performance due to fabrication process variations. MC consists in the standard approach for routine design validation~\cite{jallepalli2016employing}. On the other hand, MC based analysis requires a large number of computationally expensive circuit simulations, making it unfeasible for modern and more complex circuit technologies. To overcome this limitation, Importance sampling (IS) methods~\cite{jallepalli2016employing, shi2018fast} can be used instead to draw samples from artificially modified distributions shifted to the targeted operating zones of interest, normally towards fail regions. Thus, the convergence of the estimated robustness assessment can be accelerated since the "rare" events are prone to be drawn around these targeted regions of interest. The downside is that these methods normally favor limited specific zones of interest, therefore disfavoring the exploration and discovery of other important operating zones that end up being ignored~\cite{yin2023high}.

\medskip

{ \bf Surrogate Modeling:} Still to avoid the large number of expensive circuit simulations, data-driven surrogate models~\cite{yin2023high} can be constructed and used to approximate the circuit's response. Thanks to that, the circuit's design performance can be assessed in a faster manner. However, the main disadvantage is that these models require sufficient training samples, i.e., samples that are representative of the circuit performance dispersion. Moreover, the data-driven surrogate models suffer from the “curse of dimensionality”~\cite{yin2023high, liu2023seeking}. More specifically, the high dimensionality (which is quite common in modern analog circuits) makes it challenging to train the surrogate models. 

Fortunately, some previous research~\cite{zhai2018efficient} indicates that not all variations in input parameters are equally important. In reality, any given circuit performance measure is normally dominated by a group of critical devices, whereas the other devices have very little to no influence, especially in the case where the circuit has a symmetric design intended to alleviate the process variation~\cite{yin2023high}.

\medskip

{\bf Active Learning:} This knowledge allows to consider “dimension reduction” capable to reduce the input dimension such that only the key and relevant parameters are isolated. One possibility is to combine the efficiency of faster evaluations of surrogate models with active learning to sequentially reduce the prediction error~\cite{liu2023seeking}. \cite{yin2022efficient} adopts a Gaussian process (GP) to approximate the underlying circuit performance function and an entropy reduction criteria for active learning. One limitation is that entropy-based criterion is known for being computationally expensive. \cite{yin2023high} adds a nonlinear-correlated deep kernel method with feature selection to isolate the key determinant features to focus on and to consider in the surrogate modeling approach that also uses a GP.
These previously proposed approaches have been assessed to tackle up to 1k input dimensions, which can be considered a "small" analog circuit. A robust feature selection method is thus necessary to allow the use of surrogate modeling approaches in this application.

\medskip

{\bf Goals and contributions:} Having at our disposal an analog circuit simulator, we aim for an efficient exploration of the parameter space in order to limit the number of samples to effectively simulate. To that end, the corner stones of our method will be the tools of dimensionality reduction on the one hand, and surrogate modeling on the other hand.

\subsection{\bf Description of Sampling Flow}
\label{section:flow_description}

Within the context of integrated circuit design, the effects mentioned in Introduction are modelled by a random vector $X = \left(X^{(1)}, \dots, X^{(D)} \right)$, where each $X^{(i)}, \  i=1,\ldots,D$ is a real bounded random variable valued in $B^{(i)}$. These random variables are also called explanatory variables and $D$ is supposed to be very large. The performance of an integrated circuit is described by a random variable $Y=F(X)$, where the function $F$ is not easily accessible in practice. The function $F$ describes namely an expensive simulation and it is referred to as the "black-box" or the "computer code". Here, we shall suppose that $F : B \rightarrow \R$. Also, it will be convenient to designate, for any subset $I \subset \llbracket 1, D \rrbracket$, the associated product space as
\begin{align}
\label{eq:B_I}
    B(I) := \prod_{i \in I} B^{(i)} \ .
\end{align}
Naturally $X \in B := B(\llbracket 1, D \rrbracket) = \prod_{i=1}^D B^{(i)}$.

\medskip

\begin{figure}[htbp]
\center
\scalebox{.65}{
\begin{tikzpicture}[samplenode/.style={rectangle, draw=blue!60, fill=blue!10, thick, text width=5cm, align = center},
gsanode/.style={rectangle, draw=magenta!60, fill=magenta!10, thick, text width=4cm, align = center},
beginnode/.style={rectangle, draw=green!60, fill=green!10, thick, text width = 3cm, align = center},
endnode/.style={rectangle, draw=red!60, fill=red!10, thick, text width = 3cm, align = center},
gsanode/.style={rectangle, draw=magenta!60, fill=magenta!10, thick, text width=4cm, align = center},
squarednode/.style={rectangle, draw=yellow!60, fill=yellow!10, thick, text width=4cm, align = center},
questionnode/.style={rectangle, draw=black!60, fill=black!10, thick, text width=4cm, align = center},
invisiblenode/.style={rectangle, draw=white!60, fill=white!10, thick,minimum size=4cm, align = center},
labelnode/.style={rectangle, draw=white!60, fill=white!10, thick,text width=4cm, align = center},
short_labelnode/.style={rectangle, draw=white!60, fill=white!10, thick,text width=.5cm, align = center},
]
\node[beginnode]        (begin)       {BEGIN};
\node[samplenode]        (first_sample)       [below=of begin] {Build an initial batch of data, as Equation \eqref{eq:initial_data}, realisations of features and corresponding performances at stake. } ;
\node[gsanode]      (gsa)               [below=of first_sample]               {Use Feature Selection Algorithm \ref{alg:features_selection}, to select the $d$ features estimated as the most relevant};
\node[labelnode]    (right_label)   [below=of gsa] {$D-d$ features selected as the \textbf{less influential}};
\node[squarednode]      (right_square)       [below=of right_label] {Draw a realisation of these $D-d$ features, with the usual distribution law.};
\node[labelnode]    (left_label)   [left=of gsa] {$d$ features selected as the \textbf{most influential}};
\node[squarednode]      (left_square)       [below=of left_label] {Over the realisations for theses $d$ features, call Algorithm \ref{alg:optimized_sampling}, to obtain a new realisation of these $d$ features.};
\node[samplenode]        (new_sample)       [below=of right_square] {Consider the new realisation over all the features and call the expensive simulator on it to get the corresponding performance. Then, add the new sample to the previous ones};
\node[questionnode]		(final_budget)	[below=of new_sample] {Do we reach the final budget $N_f$ ?};
\node[short_labelnode]   (no)   [right=of final_budget]   {No};
\node[short_labelnode]   (yes)   [below=of final_budget]   {Yes};
\node[endnode]		(end)	[below=of yes] {END};

\draw[->] (begin.south) -- (first_sample.north) ;
\draw[->] (first_sample.south) -- (gsa.north);
\draw[-] (gsa.west) -- (left_label.east);
\draw[-] (gsa.south) -- (right_label.north);
\draw[->] (left_label.south) -- (left_square.north) ;
\draw[->] (right_label.south) -- (right_square.north) ;
\draw[->] (left_square.south) .. controls +(down:2cm) and +(left:3cm) .. (new_sample.west)  ;
\draw[->] (right_square.south) -- (new_sample.north);
\draw[->] (new_sample.south) -- (final_budget.north);
\draw[-] (final_budget.south) -- (yes.north);
\draw[->] (yes.south) -- (end.north);
\draw[-] (final_budget.east) -- (no.west);
\draw[->] (no.east) .. controls +(right:3cm) and +(right:5cm) .. (gsa.east);
\end{tikzpicture}
}
\caption[Diagram of our active sampling flow]
        {Diagram of our active sampling flow.}
\label{fig:general_flow}
\end{figure}
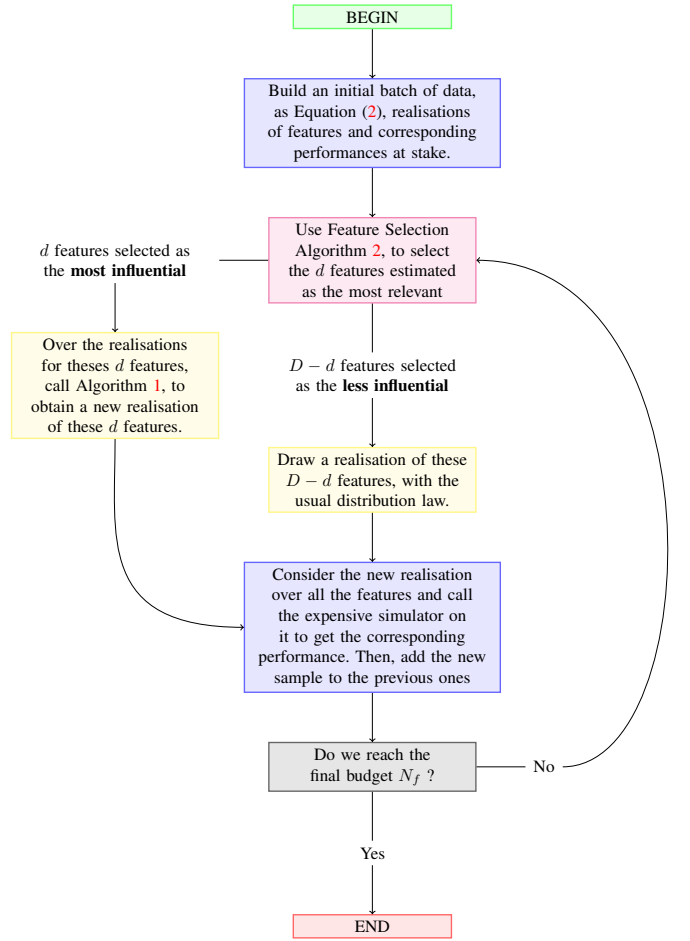

Let us now describe our active sampling flow, which is illustrated in Figure \ref{fig:general_flow}. 
We assume that we have a first batch of $N_0 \in \N^*$ realizations 
\begin{align}
\label{eq:initial_data}
    \Sc_{N_0} := \left\{ \left(x_j, y_j\right) \in B \times \R, \ 1\le j \le N_0 \right\} \ ,
\end{align}
of the random variables $\left(X, Y \right)$. The goal is to successively choose new samples, until we reached a final budget $N_f \in \N^*$, with the best characterization of circuit performance.

\medskip

{\bf Surrogate modeling: } 
Assuming we have chosen a set of samples $\Sc_N$ for $N_0 \le N < N_f$, we construct surrogate model of $F$, denoted $\widehat{\Fc}_N$ based on the observations $\Sc_N$. Then this surrogate model $\widehat{\Fc}_N$, is used in order to choose the next sample point in $B$ for which the prediction through $\widehat{\Fc}_N$ is the most uncertain. 

Among the multiple possibilities of surrogate models, we have opted for Gaussian Processes (GPs) \cite{williams2006gaussian}, for their flexibility and because of their natural Bayesian framework. Indeed, given a GP $\widehat{\Fc}_N$, that has been trained to interpolate between the values of $\Sc_N$, we have access to both predictions on the entire space
\begin{align}
\label{eq:expectation_gpr}
    \widehat{m}_N : x \mapsto \E\left[\widehat{\Fc}_N(x) \right] \ ,
\end{align}
and well as the variance
\begin{align}
\label{eq:variance_gpr}
    \widehat{v}_N : x \mapsto \Var\left[\widehat{\Fc}_N(x) \right] \ .
\end{align}

\medskip

{\bf Dimensionality reduction: }
It is well known that GPs do not scale with the dimension. Therefore, it is absolutely necessary to identify $d \ll D$ relevant indices $I_d = \left(i_1, \dots i_{d}\right) \subset \llbracket 1, D \rrbracket$ which better represent the variations at hand. Accordingly the trained Gaussian Process Regressor $\widehat{\Fc}_N$ is taken as a (random) map defined over $B(I_d)$. Likewise for $\widehat{m}_N$ and $\widehat{v}_N$.

The statistical procedure for finding $I_d$ is the Feature Selection Algorithm~\ref{alg:features_selection}, which is detailed in Section~\ref{section:features_selection} along with the necessary statistical theory.

\medskip

{\bf Optimized sampling: } 
Thus we can choose the next sample $\widetilde{x}_{N+1} \in B(I_d)$ as the one maximizing the variance, meaning that
\begin{align}
\label{eq:max_variance_criteria}
\widetilde{x}_{N+1}
= \left(\widetilde{x}_{N + 1} ^{(1)}, \dots, \widetilde{x}_{N + 1} ^{(d)} \right) = \underset{x \in  B(I_d)}{\argmax} \ \widehat{v}_N(x)\ ,
\end{align}
which we depict in Algorithm \ref{alg:optimized_sampling}. 

\begin{algorithm}
\caption{Optimized Sampling}
\label{alg:optimized_sampling}
\begin{algorithmic}
\REQUIRE{ $\Sc_N$ : the training dataset of size $N$; \par
		 \hspace{1cm} $I_d = \left(i_1,  \dots, i_{d} \right)$ : labels of relevant features; \par
          \hspace{1cm} $m$ : the number of candidate points ; \par}
\STATE{ $\widehat{\Fc}_N \leftarrow $ Gaussian Process Regressor trained on $\Sc_N$ ;}
\STATE{ $\Sc \leftarrow$ Sample $m$ i.i.d. points in $B(I_d)$ according to the law of $\left( X^{(i)},  i \in I_d \right) $ ;}
\STATE{$\widetilde{x}_{N+1} \leftarrow$ Point in $\Sc$ maximizing variance (Eq. \eqref{eq:max_variance_criteria}) ;}
\RETURN{$\widetilde{x}_{N+1} = \left( \widetilde{x}_{N+1}^{(1)}, \dots, \widetilde{x}_{N+1}^{(d)} \right) $}
\end{algorithmic}
\end{algorithm}

In order to obtain a sample $x_{N+1} \in B$, it only remains to sample the remaining (non-relevant) coordinates $z_{N+1} \in~B( \llbracket 1, D \rrbracket \backslash I_d )$. Then the new sample is obtained by concatenation $x_{N+1} := (\widetilde{x}_{N+1}, z_{N+1}) \in B$. The associated output is obtained by invoking the black-box function $F$ to compute $y_{N+1} = F\left(x_{N+1}\right)$ and finally obtain $\Sc_{N+1} := \Sc_N \cup \left\{ \left(x_{N+1}, y_{N+1} \right) \right\}$ .

\section{\bf Datasets}
\label{section:datasets}

\begin{table*}
    \centering
    \begin{tabular}{|c|ccccc|}
         \hline
         Dataset   & Sample size & Outputs / dim $Y$ & Attributes & Fictitious Attributes & Total Attributes (D)\\
         \hline
        $G_{\mathrm{Sobol'}}^{(d=4)}$ 
                   & $30 \ 000$    & $1$     & $4$ & $446$             & $450$ \\
        HSOTA      & $30 \ 000$    & $3$     & $450$ & $0$                   & $450$ \\
        $G_{\mathrm{Sobol'}}^{(d=10)}$ & $30 \ 000$    & $1$     & $10$ & $9 \ 990$             & $10 \ 000$\\
        FIRC       & $9 \ 982$     & $2$     & $44 \ 959$ & $0$                   & $44 \ 959$ \\
        \hline
    \end{tabular}
    \caption{Characteristics of datasets. 
    \textsc{The Sobol' $G$ function has extra noisy (fictitious) attributes added, in order to become comparable to the real datasets FIRC and HSOTA.}}
    \label{tab:use_cases}
\end{table*}

The datasets we use to showcase our active sampling flow are a standard Sobol G-function and simulations from actual analog circuits. Table \ref{tab:use_cases} summarizes the content of this Section. 

\medskip

\subsection{ \bf The classical Sobol' G-function}
In the community of sensitivity analysis, it is common to use the following Sobol' G-function for synthetic benchmarks \cite{azzini2022function}. 
Given $d \le D$ and $a \in \left(\R \backslash \{-1\} \right)^d$, such function $F = G_{\mathrm{Sobol'}}^{(d)}$ is defined as
\begin{align*}
\begin{array}{ll}
G_{\mathrm{Sobol'}}^{(d)}:
    & [0,1]^D \rightarrow \R \\
	& x \mapsto \prod_{i=1}^d \frac{|4x^{(i)} -2|+a_i}{1+a_i}
\end{array}   \ ,
\end{align*}
which will be only considered for $a = 0_{\R^d}$, meaning that
\begin{align}
\label{eq:g_function}
Y = G_{\mathrm{Sobol'}}^{(d)}(X) := \prod_{i=1}^d | 4X^{(i)} -2 | \ .
\end{align}

The first $d$ variables, $(X^{(1)}, \dots, X^{(d)})$, will be called the \textit{significant features}, and the $D-d$ others, $(X^{(d+1)}, \dots, X^{(D)})$, will be called the \textit{noisy features}. We add these fictitious attributes to the dataset given by the Sobol' G function, in order to artificially reproduce the same dimensionality found in the non-synthetic datasets (See Table \ref{tab:use_cases}). 

\medskip

\subsection{ \bf Simulation of analog circuits }
The analog circuits at hand are a High-Speed Operational Transconductance Amplifier (HSOTA) and a Fast Internal Reference Clock (FIRC) generator both designed in 16nm FinFET technology and composed each of few thousands of devices. See for example \cite{enwiki:1214653053, enwiki:1217842343} for general information about HSOTA amplifiers and clock circuits. We have at our disposal simulations performed thanks to SPICE-like analog simulators, which are industry standards. For example Cadence's Spectre{\textregistered} Circuit Simulator \cite{misc:spectre}. Furthermore
\begin{itemize}
    \item for the HSOTA device, the outputs are two slew rate measures (up and down) and an offset voltage, which all depend on $D = 450$ explanatory variables ;
    \item for the FIRC device, the outputs are measures of the generated clock frequency and its duty cycle, which all depend on $D = 44 \ 959$ explanatory variables.
\end{itemize}
Notice that simulations take into account circuit aging, which depends on variables describing usage scenarios. Here, we do not include such aging variables, and by explanatory variables, we only mean process and mismatch variations, temperature and supply voltages in the design of circuits.

Of course, there is no intrinsic dimension $d$ in these cases. And naturally, the chosen dimensions depend on which output is considered. In practice, dimension reduction to the values $d_{HSOTA} \in \{10, 13\}$ and $d_{FIRC} \in \{11, 15\}$ is satisfactory.

\section{\bf Features selection}
\label{section:features_selection}

\subsection{\bf Chatterjee's estimator of Cramèr-von-Mises indices}

Our method for dimensionality reduction relies on the computation of sensitivity indices, which are powerful tools from the field of Global Sensitivity Analysis, as described in \cite{gamboa2021sensitivity}. Thanks to such indices, one can assess, for $1 \le i \le D$, how sensitive is $Y$ to a variable $X^{(i)}$. Throughout the paper, we will only consider the first order Cramèr-von-Mises indices (CvM indices) which are given by 
\begin{align*}
\xi\left(X^{(i)},Y\right) 
:=  
\frac{\int_{t \in \R} \Var\left[\E\left[\mathds{1}_{Y \le t}\ | \ X^{(i)}\right]\right] \P_Y(dt) }
     {\int_{t \in \R} \Var\left[\P\left[Y\le t \right]\right] \P_Y(dt) }  \ .
\end{align*}
When there is no ambiguity, we will note $ \xi^{(i)} := \xi\left(X^{(i)},Y\right)$ and we have that $\xi^{(i)} \in [0,1]$. 

The remarkable property of the CvM index $\xi^{(i)}$ is that it vanishes if only if $Y$ and $X^{(i)}$ are independent. And it is equal to $1$ if and only if $Y$ is a function of $X^{(i)}$. As such, the CvM index $\xi^{(i)}$ quantifies the dependence, allowing us to decide whether or not  the variable $X^{(i)}$ is relevant.

We will estimate these CvM indices empirically using Chatterjee's method based on ranks. The estimator $\widehat{\xi}^{(i_0)}_N$ is defined as follows. Given a feature $1 \le i_0 \le D$, let us assume that we have at our disposal $N\in \N^*$ i.i.d. realisations of $\left(X^{(i_0)}, Y\right)$
\begin{align*}
	\Sc_N^{(i_0)} = \left\{ \left(x_j^{(i_0)}, y_j \right) \in B^{(i_0)} \times \R, \ 1 \le j \le N \right\} \ .
\end{align*}
Let us sort the sample as $\left(x_{N,(j)}^{(i_0)}, y_{N,(j)}\right)_{1 \le j \le N}$ such that
\begin{align*}
    x_{N,(1)}^{(i_0)} \le \dots \le x_{N,(N)}^{(i_0)} \ .
\end{align*}
Then we define this rank-based estimator for $\xi^{(i_0)}$ by
\begin{align}
\label{eq:xi_estimator}
    \widehat{\xi}^{(i_0)}_N := 1 - \frac{3\sum_{j=1}^{N-1} \left| r_{j+1}-r_j\right|}{N^2-1} \ ,
\end{align}
where the ranks $r_j$'s are given by
\begin{align*}
    \forall 1\le j \le N, \ r_j := \sum_{k=1}^N \mathds{1}_{ \left\{ y_{N,(k)}\le y_{N,(j)} \right\} } \ .
\end{align*}
The almost sure convergence $\lim_{N \rightarrow \infty} \widehat{\xi}^{(i_0)}_N = \xi^{(i_0)}$ is given by \cite[Theorem 1.1]{chatterjee2021new}.

Thus, because of the simplicity of this estimator, it can be incorporated into a feature selection method as Algorithm \ref{alg:features_selection}. 

\begin{algorithm}
\caption{Feature Selection}
\label{alg:features_selection}
\begin{algorithmic}
\REQUIRE{$\Sc_N$ : the training dataset of size $N$ ; \par
		\hspace{1cm} $d \in \N^*$ : the number of features to select ;}
\FOR{$i \in \llbracket 1, D \rrbracket$}
\STATE{Compute $\widehat{\xi}^{(i)}_N $ thanks to Equation \eqref{eq:xi_estimator};}
\ENDFOR
\STATE{Take the $d$ features $\left(i_1, \dots, i_{d}\right)$ maximizing $\left(\widehat{\xi}^{(i)}_N\right)_{1 \le i \le D}$}
\RETURN{$\left(i_1, \dots, i_{d} \right)$, most relevant $d$ feature labels }
\end{algorithmic}
\end{algorithm}

\subsection{\bf Conjecture for detection of noisy features}

For statistically testing whether a fixed feature is relevant, we have at our disposal the following Central Limit Theorem for one estimator $\widehat{\xi}_N^{(i)}$.

\begin{theorem}[2.2 from \cite{chatterjee2021new}]
\label{thm:clt_chatterjee}
For all $1 \le i \le D$, under the hypothesis
\begin{align*}
H_0^{(i)}: ``X^{(i)} \ \mbox{is a noisy feature} '' \ ,
\end{align*}
the corresponding Chatterjee estimator, $ \widehat{\xi}_N^{(i)} $, fluctuates according the following Central Limit Theorem (CLT)
\begin{align*}
    \sqrt{N}\widehat{\xi}_N^{(i)} \rightarrow \mathcal{N}\left(0,\frac{2}{5}\right) \ . 
\end{align*}
\end{theorem}

However, for a feature selection with a high number of noisy variables, we need a joint test, in the sense that we need to test the following hypothesis
\begin{align*}
    H_0 = H_0^{(i_1,\dots, i_{k})}: ``X^{(i_1)},\dots,  X^{(i_{k})} \ \mbox{are the noisy features'' , } 
\end{align*}
where $(i_1,\dots, i_{k}) \in \mathcal{P}_{k}\left( \llbracket 1, D \rrbracket \right)$.
In other words,
\begin{align*} 
H_0 = H_0^{(i_1,\dots, i_{k})} = H_0^{(i_1)} \cap \dots \cap H_0^{(i_{k})} \ .
\end{align*}

Thus, under such $H_0$, it is natural to track the variable
\begin{align*}
\widehat{\xi}_{\mathrm{max}}
= \widehat{\xi}_{\mathrm{max}}(N,\left( i_1, \dots i_{k} \right)) := \max_{i \in \left( i_1, \dots i_{k} \right) } \widehat{\xi}_N^{(i)} \ ,
\end{align*} 
given a sample size $N \in \N$.
Notice that maxima of random variables are not always tractable. The case that is well studied is that of the maxima of independent variables with Gaussian tails. This is the topic of classical books such as \cite{leadbetter2012extremes}. 
In fact, under $H_0$, we expect the $\widehat{\xi}_N^{(i)}$ 's to decorrelate from the samples of $Y$, so that $\widehat{\xi}_{\mathrm{max}}$ truly behaves like a maximum of i.i.d. random variables. Because Gaussian tails are a given thanks to the concentration inequalities in the proof of \cite[Theorem 1.1]{chatterjee2021new}, theorems describing extremal statistics such as \cite[Theorem 1.5.3]{leadbetter2012extremes} should still hold. Hence, we make the following conjecture
\begin{conjecture}
\label{cjt:xi_max_assumption}
Under $H_0$, we have:
\begin{align}
\label{eq:xi_max_assumption}
\widehat{\xi}_{\mathrm{max}}(N,(i_1,\dots, i_{k}))  = \sqrt{\frac{4\log(k) }{5N}} \times \gamma(N,k) \ ,
\end{align}
where the family $\left( \gamma(N,k) \ ; \ (N,k) \in (\N^*)^2 \right)$ is tight, meaning uniformly bounded with high probability .
\end{conjecture}

\begin{figure}

\centering
\includegraphics[width = .5\textwidth]{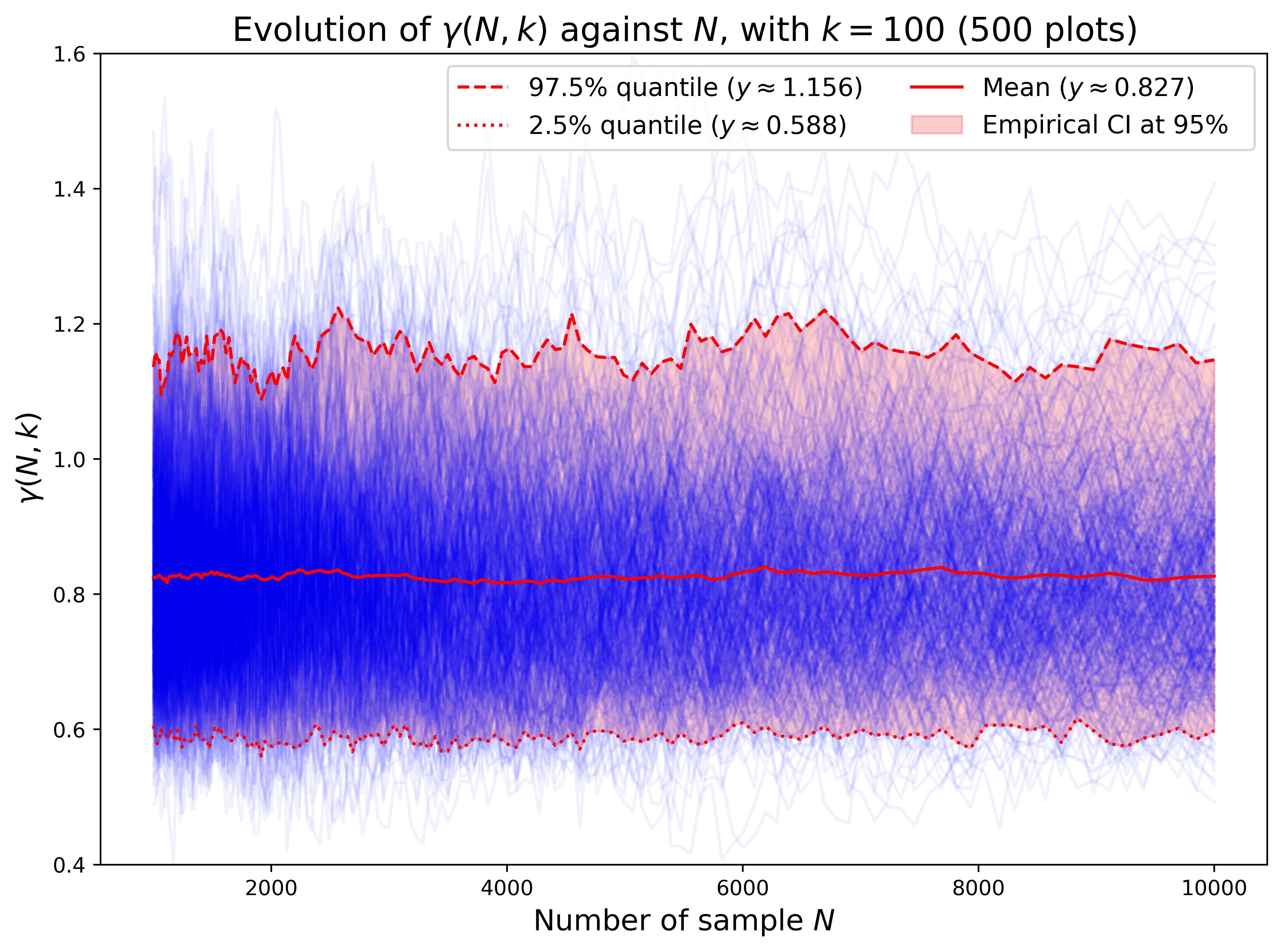}
\caption[Evolution of $\gamma(N,k)$ against $N$]{Given $N_{\mathrm{max}} = 10000$, $k=100$ and $P = 500$, we generate $N_{\mathrm{max}} \times P$ i.i.d. realisations of mutually independent random variables $\left(X^{(1)}, \dots, X^{(k)}, Y\right)$, so that we have $P$ batches of $N_{\mathrm{max}}$ samples. On each of these batches, we use the $N_{\mathrm{max}}$ samples to compute $\widehat{\xi}_{\mathrm{max}}(N,(1,\dots, k))$ for $N \le N_{\mathrm{max}}$. Thus, we obtain $P$ independent realisations of $\gamma(N, k)$ depending on $N<N_{\mathrm{max}}$, what we plot in blue.}
\label{fig:tightness_verification_N}
\end{figure}

\begin{figure}
\centering
\includegraphics[width = .5\textwidth]{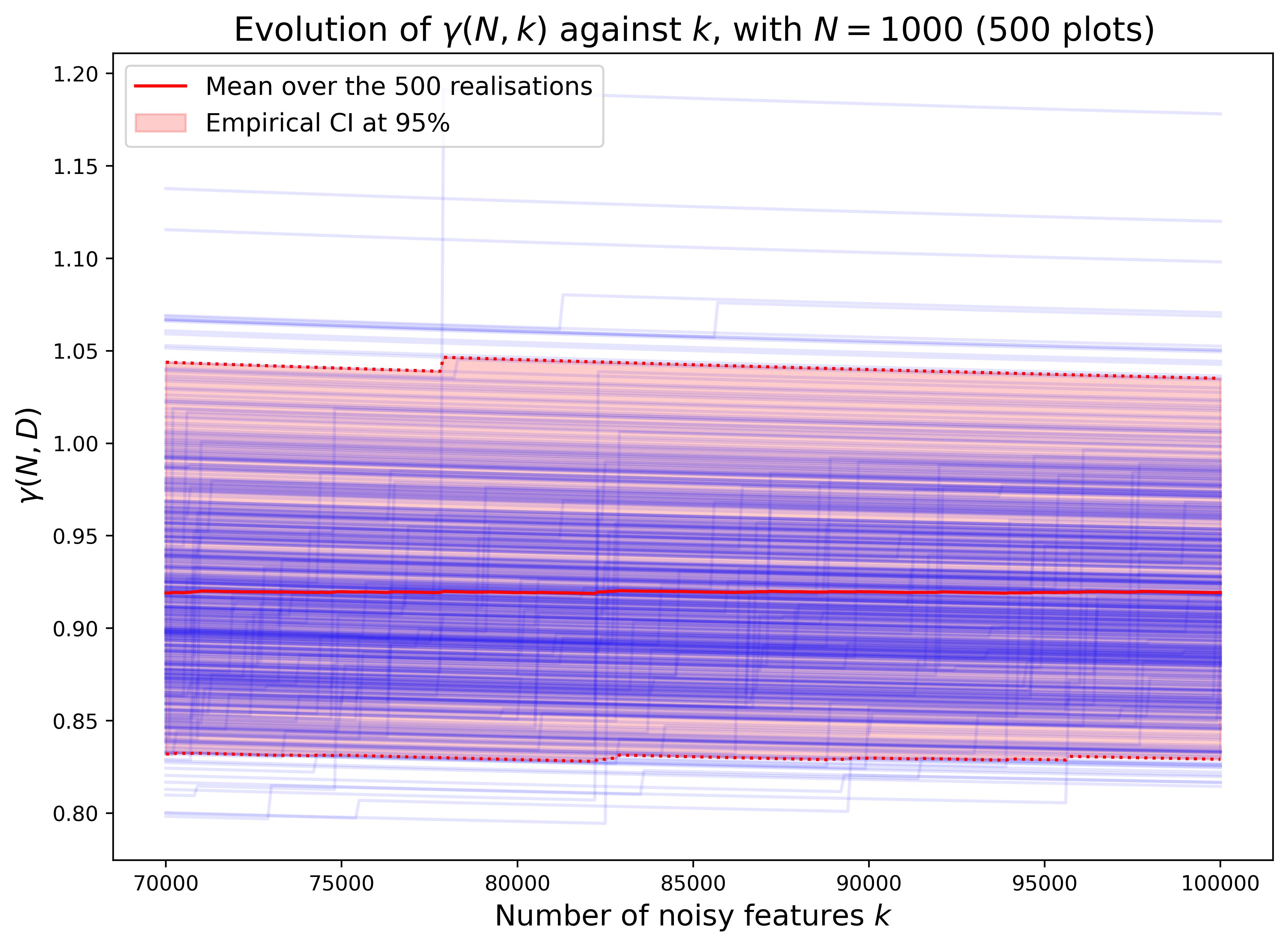}
\caption[Evolution of $\gamma(N,k)$ against $k$]{Given $N=10\ 000$, $k_{\mathrm{max}}$ and $P = 500$, we generate $N \times P$ i.i.d. realisations of mutually independent random variables $\left(X^{(1)}, \dots, X^{(k_{\mathrm{max}})}, Y\right)$, so that we have $P$ batches of $N$ samples. On each of these batches, we use the $N$ samples to compute $\widehat{\xi}_{\mathrm{max}}(N,(1,\dots, k))$ for $k \le k_{\mathrm{max}}$. Thus, we obtain $P$ independent realisations of $\gamma(N, k)$ depending on $k<k_{\mathrm{max}}$, what we plot in blue.}
\label{fig:tightness_verification_k}
\end{figure}

In order to verify empirically such a Conjecture, we start by dealing with tightness of $\gamma$ by generating a huge number of realisations of $\widehat{\xi}_{\mathrm{max}}$, which can be observed on Figures \ref{fig:tightness_verification_N} and \ref{fig:tightness_verification_k}. 

We also make some additional numerical experiments to test the relevance of Equation \eqref{eq:xi_max_assumption}. Indeed, taking the logarithm, we obtain
\begin{align}
\label{eq:log_xi_max_assumption}
  & \log \widehat{\xi}_{\mathrm{max}}(N,(i_1,\dots, i_{k})) \\
= & - \alpha_1 \log \frac{5N}{2} + \beta_1 \log \left(2 \log(k)\right) 
    + \log \gamma(N,k) \ , \nonumber
\end{align}
where 
$$ \alpha_1 = \beta_1 = \half \ .$$

\begin{figure}
\centering
\includegraphics[width = .5\textwidth]{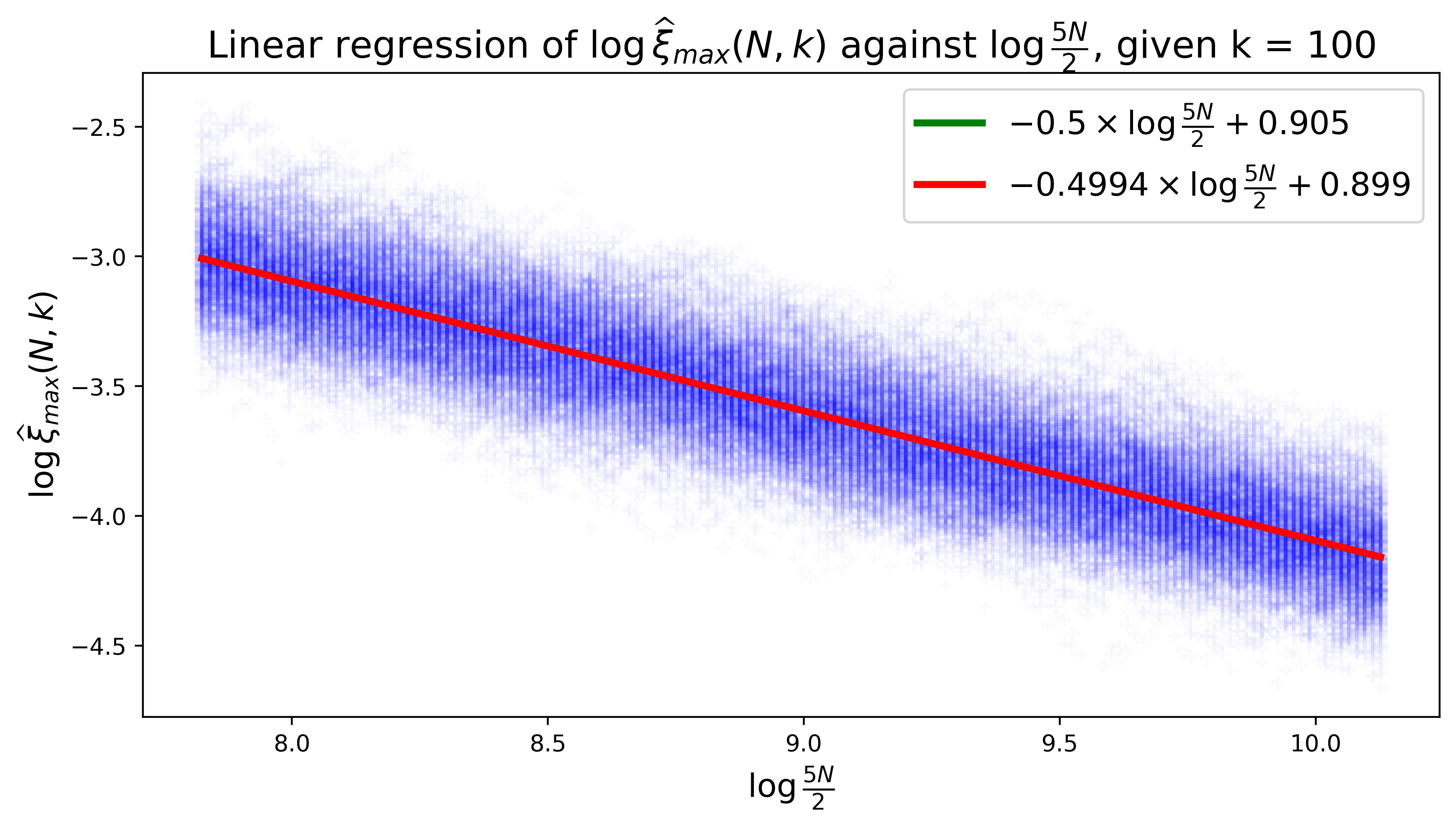}
\caption[Evolution of $\log(\widehat{\xi}_{\mathrm{max}}(N,\left( i_1, \dots i_{k} \right)))$ against $\log\left(\frac{5N}{2}\right)$]{We perform an (OLS) linear regression to estimate  the slope $-\alpha_1$ in Equation \eqref{eq:log_xi_max_assumption}. We obtain a confidence interval for $\alpha_1$ of $[0.498 \ , \ 0.501 ]$ with risk level at $95\%$.}
\label{fig:regression_N}
\end{figure}

\begin{figure}
\centering
\includegraphics[width = .5\textwidth]{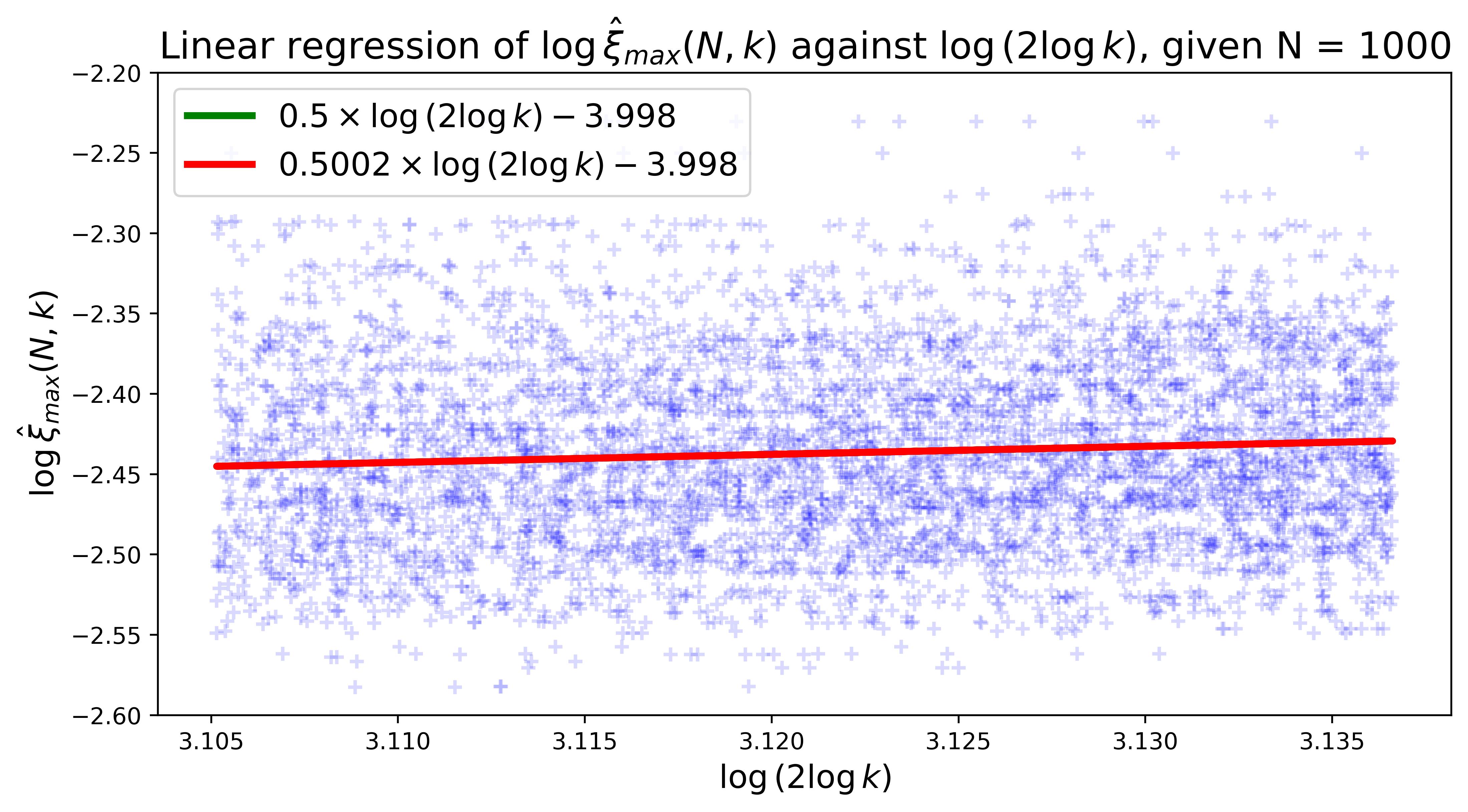}
\caption[Evolution of $\log(\widehat{\xi}_{\mathrm{max}}(N,\left( i_1, \dots i_{k} \right)))$ against $\log \left(2 \log(k) \right)$]{We perform an (OLS) linear regression to estimate  the slope $\beta_1$ in Equation \eqref{eq:log_xi_max_assumption}. We obtain a confidence interval of $[0.497 \ ; \ 0.504]$  with risk level at $95\%$.}
\label{fig:regression_k}
\end{figure}

As $\log \gamma(N,k)$ is bounded with high probability, it can be treated as a residue. And the values of $\alpha_1$ and $\beta_1$ can be tested thanks to a classical Ordinary Least Squares (OLS) regression. The results in Figures \ref{fig:regression_N} and \ref{fig:regression_k} tend to confirm the conjecture as well.

\section{\bf Experiments}
\label{section:experiment_validation}

{\bf Variations of the flow: } In order to assess the quality of our active sampling flow as described in Fig. \ref{fig:general_flow}, we consider different variations, in the feature selection and the sample choice. The different methods are listed in Table \ref{tab:flow_stragegies} and are motivated as follows.
\begin{table}
\centering
    \begin{tabular}{|c|c|c|}
        \hline
        & \textbf{Feature selection} & \textbf{Choice of next sample} \\
        \hline
        \textit{Method 1} & GSA (Algorithm \ref{alg:features_selection}) & Maximal variance (Algorithm \ref{alg:optimized_sampling}) \\ 
        \hline
        \textit{Method 2} & GSA (Algorithm \ref{alg:features_selection}) & Random Choice \\
        \hline
        \textit{Method 3} & Oracle Selection & Maximal variance (Algorithm \ref{alg:optimized_sampling})  \\ 
        \hline
        \textit{Method 4} & Oracle Selection & Random Choice \\ 
        \hline
        \textit{Method 5} & Random Selection & Maximal variance (Algorithm \ref{alg:optimized_sampling})  \\ 
        \hline
        \textit{Method 6} & Random Selection & Random Choice \\ 
        \hline
    \end{tabular}
    \caption{Variations of the active sampling flow of Fig. \ref{fig:general_flow} during our experiments.}
    \label{tab:flow_stragegies}
\end{table}

In the spirit of ablation studies, the choice of next sample can be done using the maximal variance criterion (Algorithm \ref{alg:optimized_sampling}) or using a random choice. Regarding the feature selection, more variations are possible. A priori, the worst choice is a random selection. Then there is our feature selection based on sensitivity analysis (Algorithm \ref{alg:features_selection}). And finally there is the ideal case where we know of most relevant features thanks to an oracle.

For synthetic dataset obtained from the Sobol' G-function, the oracle is simply the knowledge of the truly relevant variables during the dataset's generation. For the HSOTA and the FIRC, the oracle is given by a sensitivity analysis performed on the entire dataset before launching the exploration.

\medskip

{\bf Design of experiment and performance metric:} 
We first start by separating our datasets into a training subset and a testing subset, according to a 75\%-25\% split ratio. Then, we perform the same following procedure for each method in Table \ref{tab:flow_stragegies}. 

For any $N_0 \le N \le N_f$, the (variant of the) active sampling flow provides a sample set
\begin{align*}
    \Sc_N = \left\{ \left(x_j, y_j\right) \in B \times \R, \ 1\le j \le N \right\} \ ,
\end{align*}
which is a subset of the training dataset. The sample $\Sc_N$ is used to train a Gaussian Process Regressor $\widehat{\Fc}_N$ over the selected features. 

We are thus in position of computing an $R^2$ score on the testing dataset $\left( x^{test}_j, y^{test}_j \right)_{1\le j \le N_T}$. For convenience, we recall the definition of the $R^2$ coefficient, also known as determination score
\begin{align*}
    R^{2}(N) = 1 - 
    \frac{ \sum_{j=1}^{N_{T}}\left( y^{\mathrm{test}}_j
                                 -  \E\left[ \Fc_{N}\left(x^{\mathrm{test}}_j\right) \right] 
                                 \right)^2 }
    {   \sum_{j=1}^{N_{T}}\left(y^{\mathrm{test}}_j - \overline{y} \right)^2
    } \ ,
\end{align*}
where $\overline{y}$ is the mean
\begin{align*}
   \overline{y} = \frac{1}{N_T}\sum_{j=1}^{N_T} y^{\mathrm{test}}_j \ .
\end{align*}

\begin{figure}
\centering
\includegraphics[width=.5\textwidth]{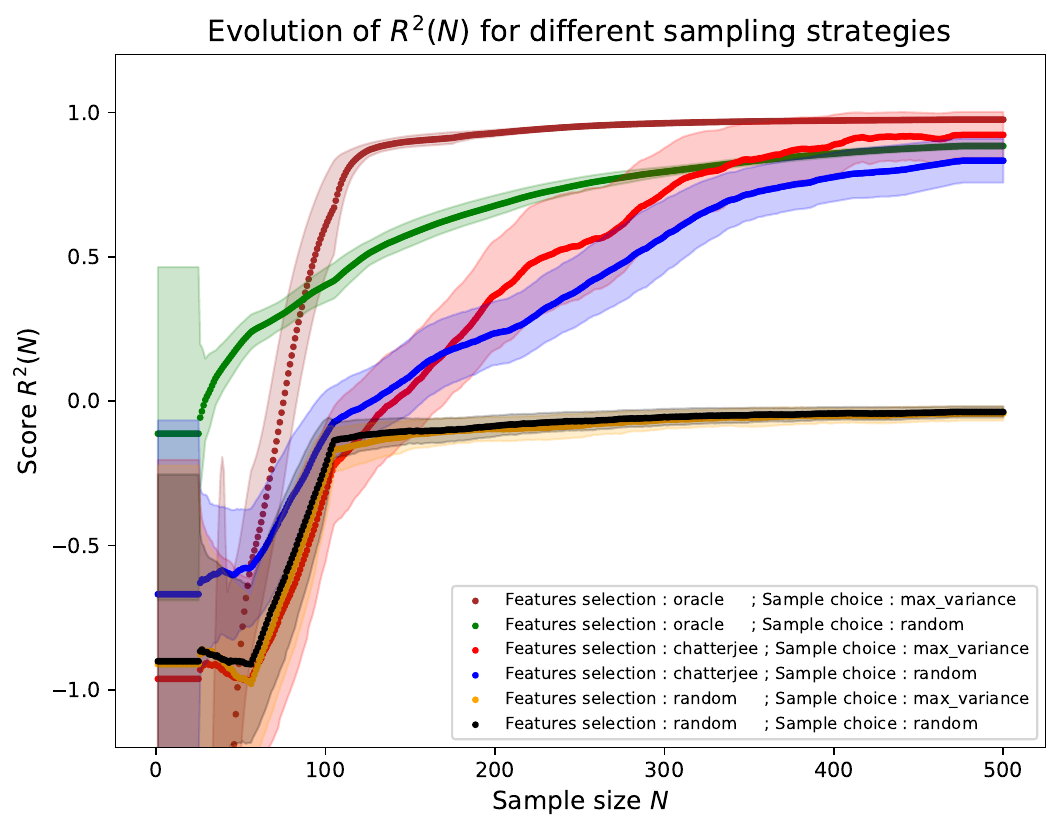}
\caption[Evolution of $R^{2}(N)$ depending on each methods described by Table \ref{tab:flow_stragegies} for G-function]{We compute the mean  $R^{2}(N)$ over $20$ runs for each method, as $N$ grows, as described through Section \ref{section:experiment_validation}, for the dataset built from $G_{\mathrm{Sobol'}}^{d=4}$.}
\label{fig:r2_mean_g4}
\end{figure}

In Figure \ref{fig:r2_mean_g4}, we illustrate the result graphically in the case of $F= G_{\mathrm{Sobol'}}^{(d=4)}$. A more comprehensive account of the results will be given in a table.

\medskip

{\bf Discussion:} It seems that our flow (Method 1 - The original flow of Fig. \ref{fig:general_flow}) outperforms all other sampling methods, except for the Method 3, which is best case scenario using oracle a priori information.

\section{\bf Conclusion}
\label{section:conclusion}

In this paper, we have shown the relevance of the statistical tools from sensitivity analysis, in order to design an active sampling flow for the characterization of analog circuits' performances against a large number of variables representing sources of parametric variation. This active sampling flow outperforms Monte-Carlo sampling and we have illustrated the role of the different ingredients.

The framework is flexible enough to include surrogate models other than GPs, or to seek sampling criteria other than maximum variance. As such, one could include expert knowledge in the surrogate models or one could optimize for other criteria such as failure rates. This could be the topic of future research.


\bibliographystyle{ieeetr}
\bibliography{biblio.bib}

\end{document}


\title{ Sensitivity Analysis for Active Sampling, with Applications to the Simulation of Analog Circuits
\thanks{The authors R.C., F.G. and C.P acknowledge the support of the ANR-3IA ANITI (Artificial and Natural Intelligence Toulouse Institute).}
}

\author{\IEEEauthorblockN{CHHAIBI Reda}
\IEEEauthorblockA{\textit{Université Toulouse III - Paul Sabatier} \\
Toulouse, France \\
reda.chhaibi@math.univ-toulouse.fr \\
ORCID: 0000-0002-0085-0086}
\and
\IEEEauthorblockN{GAMBOA Fabrice}
\IEEEauthorblockA{\textit{Université Toulouse III - Paul Sabatier} \\
Toulouse, France \\
fabrice.gamboa@math.univ-toulouse.fr\\
ORCID: 0000-0001-9779-4393}
\and
\IEEEauthorblockN{OGER Christophe}
\IEEEauthorblockA{\textit{NXP Semiconductors} \\
Toulouse, France \\
christophe.oger@nxp.com}
\and
\IEEEauthorblockN{OLIVEIRA Vinicius}
\IEEEauthorblockA{\textit{NXP Semiconductors} \\
Toulouse, France \\
vinicius.alvesdeoliveira@nxp.com}
\and
\IEEEauthorblockN{PELLEGRINI Clément}
\IEEEauthorblockA{\textit{Université Toulouse III - Paul Sabatier} \\
Toulouse, France \\
clement.pellegrini@math.univ-toulouse.fr\\
ORCID: 0000-0001-8072-4284}
\and
\IEEEauthorblockN{REMOT Damien}
\IEEEauthorblockA{\textit{Université Toulouse III - Paul Sabatier}\\
Toulouse, France \\
damien.remot@math.univ-toulouse.fr}
\and
Authors are in alphabetical order.
}

\maketitle

\thispagestyle{plain}
\pagestyle{plain}



\medskip

\bibliographystyle{ieeetr}
\bibliography{biblio.bib}

\newpage

\appendix

\begin{algorithm}
\caption{Active Sampling Flow}
\label{alg:general_flow}
\begin{algorithmic}
\REQUIRE{$\Sc_{N_0}$ : the initial training dataset of size $N_0$; \par
         \hspace{1cm} $N_f \ge N_0$ : the final sample budget ;\par
         \hspace{1cm} $m$ : the number of candidate samples at each step \par
         \hspace{1cm} $F: B \rightarrow \R$  : the expensive and real simulator ; \par
         \hspace{1cm} $d$ : number of features selected to build the GP ;
         }
         
\STATE{ Initialisation:  $N \leftarrow N_0$ ;}
\WHILE{$N < N_f$}
\STATE{$\left(i_1, \dots, i_{d} \right) \leftarrow$ Call Feature Selection Algorithm \ref{alg:features_selection} with inputs $\Sc_N$ and $d$  ;}
 \STATE{$\widetilde{x}_{N+1} \leftarrow $ Call Optimized Sampling Algorithm \ref{alg:optimized_sampling}, with inputs $\Sc_N$, $\left(i_1, \dots, i_{d} \right)$, $m$;}
 \STATE{Take a new independent sample $z_{N+1} \in B$ from the law of $X$ ;}
 \STATE{Build $x_{N+1} \in B$ from $\widetilde{x}_{N+1}$ and $z_{N+1}$ as explained by Equation \eqref{eq:max_variance_criteria}. }
 \STATE{$y_{N+1} \leftarrow F\left(x_{N+1}\right)$ ;}
 \STATE{$\Sc_{N+1} \leftarrow \Sc_N \cup \left\{ \left(x_{N+1}, y_{N+1} \right) \right\}$ ; }
 \STATE{$N \leftarrow N + 1$;}
 \ENDWHILE
\RETURN{ $S_{N_f}$ }
\end{algorithmic}
\end{algorithm}

\subsection{Gaussian Process Regression}
\label{appendix:gaussian_process_regression}
We recall some notions about the construction of a model regression using Gaussian Process. First, we assume that we have at our disposal $N$ i.i.d. realisations of $X$~:
\begin{align*}
\forall 1\le j \le N, \ x_j = \left(x^{(1)}_j, \dots, x^{(D)}\right) \in B \ .
\end{align*}
We also have the corresponding output $y\in \R^N$, as specified by Equation \eqref{eq:black_box}~:
\begin{align*}
\forall 1 \le j \le N, \ y_j = F\left(x_j^{(1)}, \dots, x_j^{D}\right) \ .
\end{align*}

As explained in Section \ref{sec:preliminairies}, we identify $d$ relevant features $\left(i_1, \dots i_{d}\right)$ among the $D$ total features to not suffer from the dimensionality of the simulation. To ease the notation, we consider for this section, that $F$ is defined on $\widetilde{B} \subset \R^d$ defined by Equation \eqref{eq:b_tilde}, meaning that $F$ only depends on these more relevant features, labelled by $\left(i_1, \dots i_{\tilde{d}}\right)$.

Given a kernel $K:\R^d \times \R^d \rightarrow \R$, we consider the following Gaussian Process over $\R^d$ :
\begin{align*}
\Fc \sim \Gc\Pc\left(0, K\right) \ ,
\end{align*}
where $0$ referees to the constant function over $\R^d$, equals to $0$. Now let us consider the Gaussian vector~:
\begin{align*}
\widehat{Y} = \left(\begin{array}{c}
		\Fc\left(x_1\right) \\
		\dots \\
		\Fc\left(x_N\right) 
\end{array} \right) \sim \Nc\left(0_{\R^{N}}, K_x \right) \ ,
\end{align*}
where :
\begin{align*}
	K_x := \left(K\left(x_j, x_k \right)\right)_{1\le j, k \le N} \in \Mc_{N}(\R) \ .
\end{align*} 

When observations $y$ are pre-processed to be normalized, what is often done in practice, then the expectation of $\widehat{Y}$ matched with the empirical mean of $y$. Now, for a point $z\in \widetilde{B}$, we want to make a prediction for $F(z)$, that is why we look at $\Fc(z)$, conditionally to $\widehat{Y}=y$. Thus, using natural properties about Gaussian Process~:
\begin{align*}
\left(\begin{array}{c}
		\widehat{Y} \\
		\Fc\left(z\right)
\end{array} \right) \sim \Nc\left(0_{\R^{N+1}},
\left(\begin{array}{cc}
		K_x & k(z) \\
		k(z)^t & K(z,z) 
\end{array}\right) \right) \ ,
\end{align*} 
where :
$$k(z) := \left(\begin{array}{c}
		K\left(x_1, z\right) \\
		\dots \\
		K\left(x_{N}, z\right) 
\end{array}\right)\in \R^{N} \ , $$
and thus, we get the new Gaussian Process~:
\begin{align*}
& \widehat{\Fc}_N = \left(\Fc(z) \ | \ \widehat{Y} = y \right) \\
& \ \sim \Nc\left(\langle k(z), \  K_{x}^{-1}y \rangle ,\ K(z,z)-\langle k(z), \ K_x^{-1}k(z) \rangle \right) \ .
\end{align*}

Then, we can take the expectation of this Gaussian Process as estimator of $F(z)$. It defines a model $\widehat{F}_N$~:
\begin{align}
\label{eq:expectation_gpr_detailled}
    \widehat{F}_N : z \mapsto  \E\left[\Fc\left(z\right) \ | \ \widehat{Y} = y \right] =  \langle k(z), \  K_{x}^{-1}y \rangle \ .
\end{align}
We can also notice that it also gives us an explicit formula for the variance~:
\begin{align*}
    \Var\left[\Fc(z) \ | \ \widehat{Y} = y \right] =  K(z,z)-\langle k(z), \  K_x^{-1}k(z) \rangle \ ,
\end{align*}
which gives us an information about the uncertainties of the prediction at points $z\in B$. Now, we can make explicit Equation \eqref{eq:max_variance_criteria}~:
\begin{align*}
    \widetilde{x}_{N+1} & = \argmax_{z \in \widetilde{B} } \ \Var\left[\Fc(z) \ | \ \widehat{Y} = y \right] \\
    & = \argmax_{z \in \widetilde{B} } \  \left( K(z,z)-\langle k(z), \  K_x^{-1}k(z) \rangle \right) \ .
\end{align*}

Also, to make our strategy of active sampling faster, we try to add not only $1$ sample but a batch of $n$ samples at each step. Our first idea was to choose a large set of points in $\widetilde{B}$ (chosen according to the initial sampling law for instance) and to take the $n$ points maximizing 
\begin{align}
\label{eq:multi_max_variance_criteria}
    z \mapsto \Var\left[\Fc(z) \ | \ \widehat{Y} = y \right]\ .
\end{align}

\todo{! Warning ! Issue to handle~: Choice of the next samples !}

Nevertheless, doing that way, we take the risk to add useless samples, in the sense where, if some of the selected samples were maximizing the same local maximum, then we probably need just one of them to already have a good reduction of the variance around such local maximum.

Finally, we can notice that, for any $1\le j \le N$, we have~:
\begin{align*}
    \widehat{F}_N\left(x_j\right) = \E\left[\Fc\left(x_j\right) \ | \ \widehat{Y} = y \right] = y_j \ , \\
    \Var\left[\widehat{\Fc}_N(x_j)\right] = \Var \left[\Fc\left(x_j\right) \ | \ \widehat{Y} = y \right] =  0 \ .
\end{align*}

\subsection{FOCI}

\todo{Reda: Pourquoi parler de FOCI dans le corps? Et les infos qu'il y a ici doivent être bcp plus fines pour comprendre ce qui se passe. }

We used a feature selection algorithm called Feature Ordering by Conditional Dependence (FOCI), provided by \cite{azadkia2021simple}, in order to identify features that are the most relevant to explain all the 5 outputs described up there:
\begin{itemize}
    \item for slew up rate from HSOTA, we identify a subset of $10$ relevant features among the $450$ ;
    \item for slew down rate from HSOTA, we identify a subset of $10$ relevant features among the $450$ ;
    \item for offset voltage from HSOTA, we identify a subset of $13$ relevant features among the $450$ ;
    \item for outputs from FIRC, we have for now memory issues, which prevent us to deal with the $9 \ 982$ samples of the $44 \ 959$ features through FOCI algorithm.
\end{itemize}

\subsection{Complements for Conjecture \ref{cjt:xi_max_assumption}}

\begin{theorem}[1.5.3 from \cite{leadbetter2012extremes}]
\label{thma:extremal_cvg}
Let $\left(U^{(k)}\right)_{k \in \N^*}$ be an i.i.d standard Gaussian random sequence. Then~:
\begin{align*}
a_k\left(M_k-b_k\right) \underset{k \to +\infty}{\overset{\Lc}{\longrightarrow}} \Gc \ .
\end{align*}
where
\begin{align*}
    M_k &= \max_{1 \le i \le k} U^{(i)} \ , \\
    a_k &= \left(2 \log k\right)^{1/2} \ , \\ 
    b_k &= \left(2 \log k\right)^{1/2} - \half \left(2 \log k \right)^{-1/2}\left(\log 4\pi + \log \log k \right) \ ,
\end{align*}
and where $\Gc$ is the standard Gumbel law, characterized by the following cumulative distribution function
\begin{align*}
    F_{\Gc}~: x \mapsto \exp\left(-e^{-x}\right) \ . 
\end{align*}
\end{theorem}

Here is the theorem for about maxima of independent Gaussian variable, which gives the motivation for the Conjecture \ref{cjt:xi_max_assumption}.

\subsection{Validation of GP as surrogate model}

In this section, we want to verify whether the proposed active learning flow using the criteria given in Equation \eqref{eq:max_variance_criteria} is relevant or not. 

Given an initial budget, $N_0$, and a final budget $N_f$, we compare two ways of sampling our data, given by Algorithms \ref{alg:active_flow} and \ref{alg:random_flow}. We can notice that Algorithm \ref{alg:active_flow} is Algorithm \ref{alg:general_flow}, with $d=D$ and $\left(i_1, \dots, i_d\right) = \left(1, \dots, D\right)$. Respectively, let us call $\Sc_{N_f}^{\mathrm{active}}$ and $\Sc_{N_f}^{\mathrm{random}}$ the results of both algorithms.

\begin{algorithm}
\caption{Active Sampling Flow (low dimension)}
\label{alg:active_flow}
\begin{algorithmic}
\REQUIRE{$\Sc_{N_0}$ : the initial training dataset of size $N_0$; \par
         \hspace{1cm} $N_f \ge N_0$ : the final sample budget ;\par
         \hspace{1cm} $m$ : the number of candidate samples at each step ; \par
         \hspace{1cm} $F~: B \rightarrow \R$ : the expensive simulator ; \par
         }
         
\STATE{ Initialisation:  $N \leftarrow N_0$ ;}
\WHILE{$N < N_f$}
 \STATE{Call Optimized Sampling Algorithm \ref{alg:optimized_sampling}, requiring  $\Sc_N$, $\left(1, \dots, D\right)$, $m$,  ensuring $x_{N+1}$;}
 \STATE{$y_{N+1} \leftarrow F\left(x_{N+1}\right)$ ;}
 \STATE{$\Sc_{N+1} \leftarrow \Sc_N \cup \left\{ \left(x_{N+1}, y_{N+1} \right) \right\}$ ; }
 \STATE{$N \leftarrow N + 1$;}
 \ENDWHILE
\RETURN{ $S_{N_f}$ }
\end{algorithmic}
\end{algorithm}

\begin{algorithm}
\caption{Random Sampling Flow}
\label{alg:random_flow}
\begin{algorithmic}
\REQUIRE{$\Sc_{N_0}$ : the initial training dataset of size $N_0$; \par
         \hspace{1cm} $N_f \ge N_0$ : the final sample budget ;\par
         \hspace{1cm} $F~: B \rightarrow \R$ : the expensive simulator ; \par
         }
         
\STATE{ Initialisation:  $N \leftarrow N_0$ ;}
\WHILE{$N < N_f$}
 \STATE{Take a new independent sample $x_{N+1} \in B$ from the law of $X$ ;}
 \STATE{$y_{N+1} \leftarrow F\left(x_{N+1}\right)$ ;}
 \STATE{$\Sc_{N+1} \leftarrow \Sc_N \cup \left\{ \left(x_{N+1}, y_{N+1} \right) \right\}$ ; }
 \STATE{$N \leftarrow N + 1$;}
 \ENDWHILE
\RETURN{ $S_{N_f}$ }
\end{algorithmic}
\end{algorithm}

Then, for each $N_0 \le N \le N_f$, we use $\Sc_{N}^{\mathrm{active}}$ and $\Sc_{N}^{\mathrm{random}}$ to build respectively two different Gaussian Process Regressors~: $\widehat{\Fc}_N^{\mathrm{active}}$ and $\widehat{\Fc}_N^{\mathrm{random}}$. Then, we look at their expectations, similarly at Equation \eqref{eq:expectation_gpr_detailled}, to respectively obtain following models~: $\widehat{F}_N^{\mathrm{active}}$ and $\widehat{F}_N^{\mathrm{random}}$ . Finally, given a collection of $N_T \in \N^*$ independent realisations $\left(x^{test}_j, y^{test}_j\right)_{1\le j \le N_T}$ of $\left(X,Y\right)$, used as testing set, we compute the corresponding score of determination, for each  $\mathrm{method} \in \left\{ \mathrm{active}, \mathrm{random} \right\}$~:
\begin{align*}
    R^{2}_{\mathrm{method}}(N) = 1 - \frac{\sum_{j=1}^{N_{T}}\left( y^{\mathrm{test}}_j - \widehat{F}^{\mathrm{method}}_{N}\left(x^{\mathrm{test}}_j\right)\right)^2 }{\sum_{j=1}^{N_{T}}\left(y^{\mathrm{test}}_j - \overline{y} \right)^2} \ ,
\end{align*}
where~:
\begin{align*}
   \overline{y} = \frac{1}{N_T}\sum_{j=1}^{N_T} y^{\mathrm{test}}_j \ .
\end{align*}

\subsubsection{Active Sampling VS Random Sampling for G-function}
\label{appendix:low_dim_g4_exp}

We study such metric with the following framework :
\begin{itemize}
\item $N_0 = 1$ and $N_f = 200$ ;
\item the expensive simulator is $F = G^{d(=4)}$ and $D=4$ ;
\item for Algorithm \ref{alg:active_flow}, $m=10^5$ ;
\item the number of data in testing set is taken at $N_T = 10^3$.
\item for fixed $N_0 \le N \le N_f$ and $\mathrm{method} \in \left\{ \mathrm{active}, \mathrm{random} \right\}$, we made $P=50$ independent realisations of $R^{2}_{\mathrm{method}}(N)$, by making computations from $P$ different and independent batches of $S_{N_0}$.
\end{itemize}

\begin{figure}
\centering
\includegraphics[width=.5\textwidth]{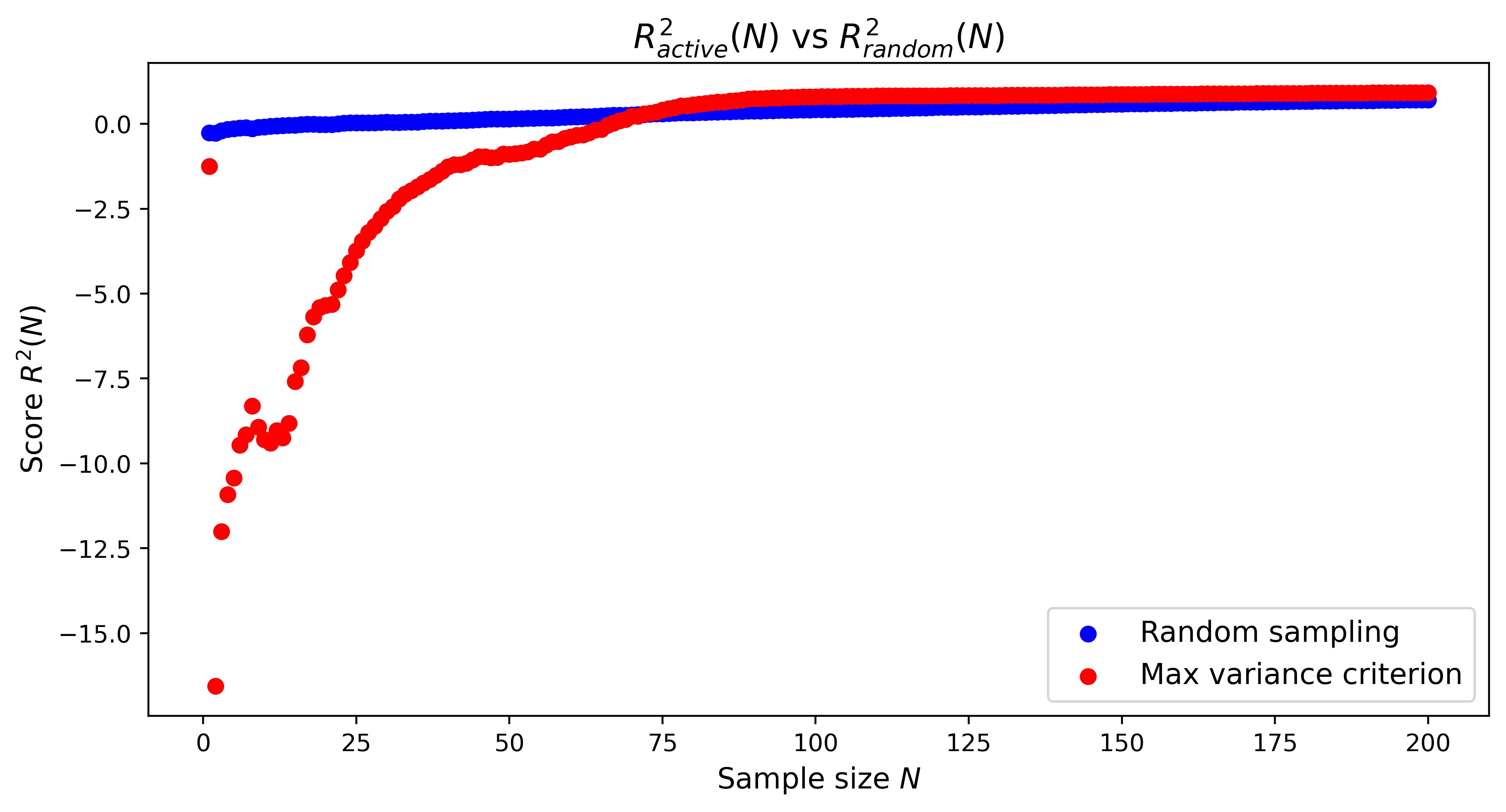}
\caption[Evolution of $R^{2}_{\mathrm{active}}(N)$ vs $R^{2}_{\mathrm{random}}(N)$ for G-function]{We compute the mean over the $P=50$ realisations of $R^{2}_{\mathrm{active}}(N)$ (in red) and $R^{2}_{\mathrm{random}}(N)$ (in blue) against $N$, as described through the framework of Section \ref{appendix:low_dim_g4_exp}).}
\label{fig:r2_mean_g4_low_dim}
\end{figure}

\begin{figure}
\centering
\includegraphics[width=.5\textwidth]{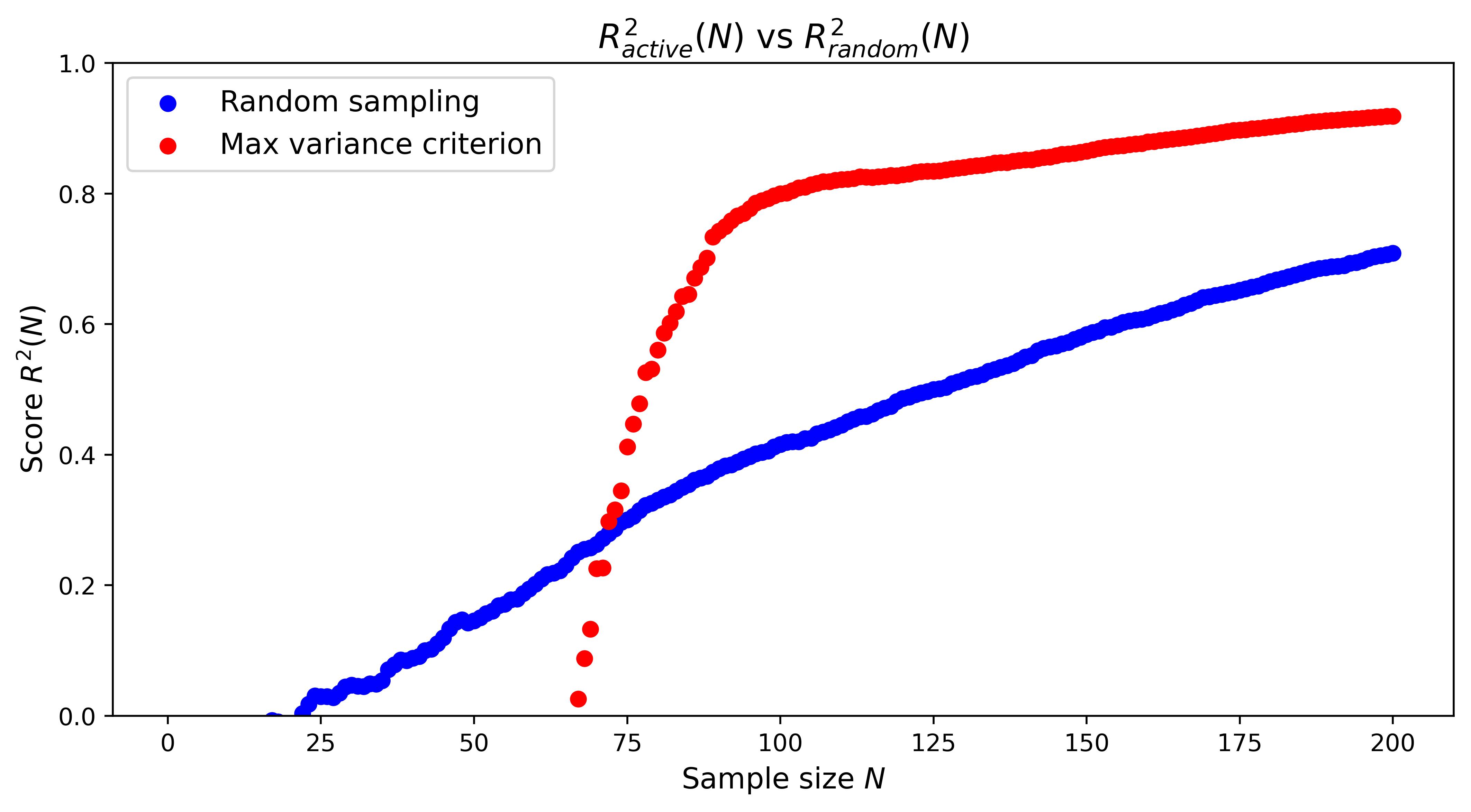}
\caption[Evolution of $R^{2}_{\mathrm{active}}(N)$ vs $R^{2}_{\mathrm{random}}(N)$ for G-function (scaled)]{Same as Figure \ref{fig:r2_mean_g4_low_dim}, but with y-axis scaled between $0$ and $1$.}
\label{fig:r2_mean_g4_low_dim_scaled}
\end{figure}

\begin{figure}
\centering
\includegraphics[width=.5\textwidth]{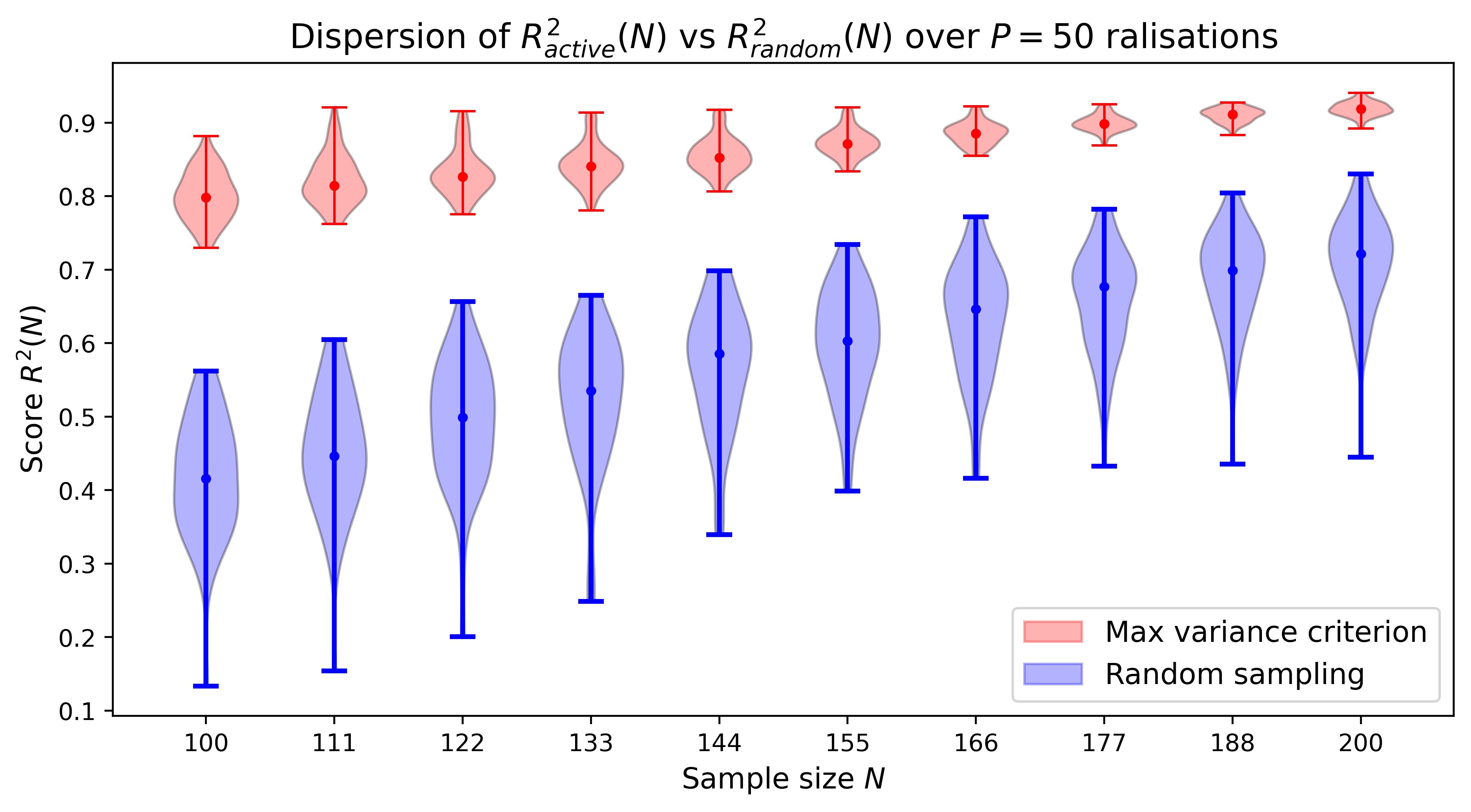}
\caption[Dispersion of $R^{2}_{\mathrm{active}}(N)$ vs $R^{2}_{\mathrm{random}}(N)$ for G-function]{We made the violin plots over the $P=50$ realisations of $R^{2}_{\mathrm{active}}(N)$ (in red) and $R^{2}_{\mathrm{random}}(N)$ (in blue) against $N$ that we generated through experiments described in Section \ref{appendix:low_dim_g4_exp}).}
\label{fig:r2_violinplot_g4_low_dim}
\end{figure}

On Figure \ref{fig:r2_mean_g4_low_dim} (scaled version available on Figure \ref{fig:r2_mean_g4_low_dim_scaled}), we can observe the mean behavior of $R^2_{\mathrm{active}}(N)$ and $R^2_{\mathrm{random}}(N)$ through the $P=50$ realizations and on Figure \ref{fig:r2_violinplot_g4_low_dim}, we use violin plots to see the distribution of these scores through the $P=50$ realizations. It seems that our flow outperforms random sampling for high sample size, nevertheless, we can observe that for small sample size, our active sampling flow is worse than the random sampling flow. \\

\subsubsection{Hybrid Sampling for G-function}
\label{appendix:hybrid_g4_exp}

\begin{figure}
\centering
\includegraphics[width=.5\textwidth]{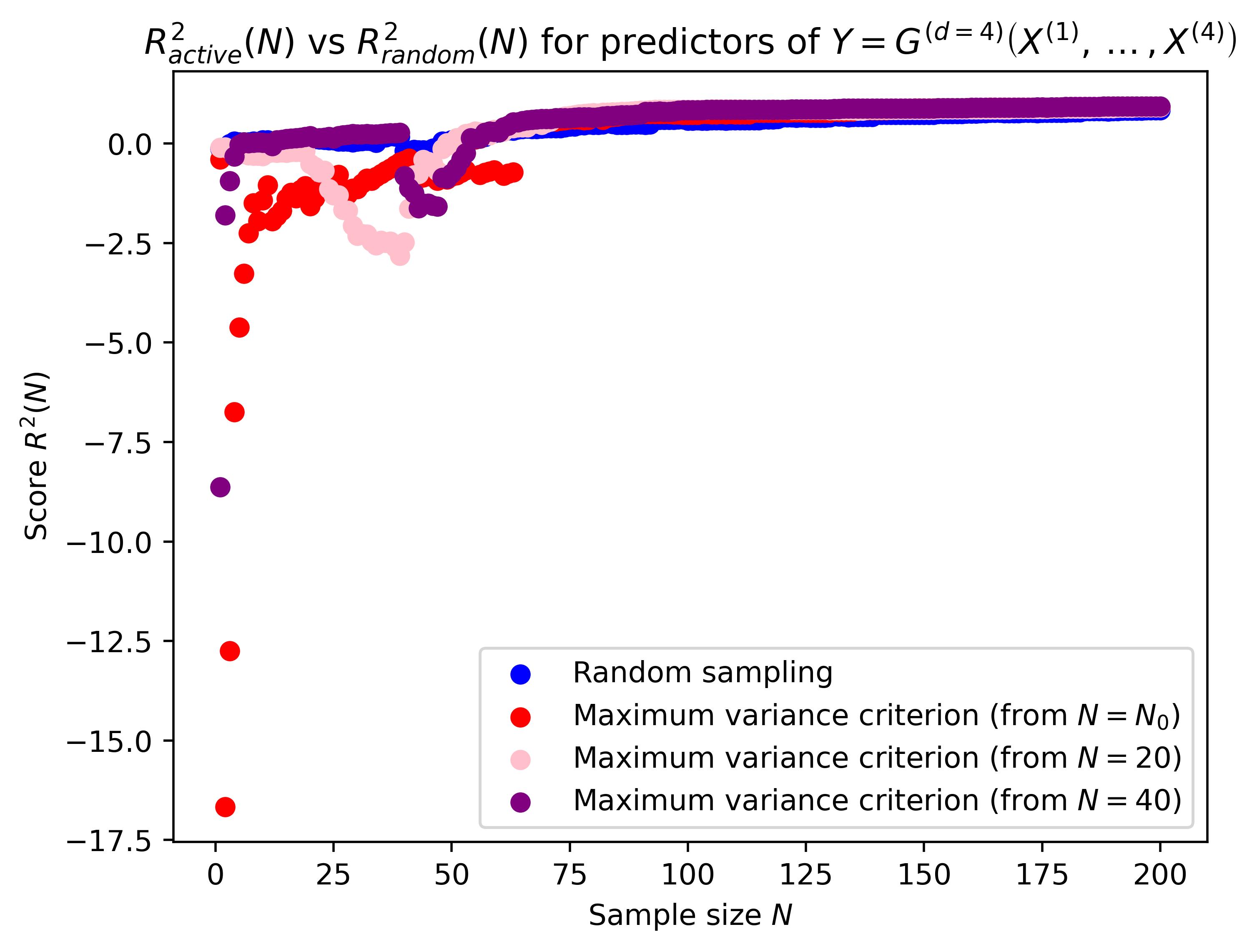}
\caption[$R^{2}(N)$ depending on some samplings methods for G-function]{We compare sampling methods from $N_0 = 1$ until $N_f = 200$: active sampling from Algorithm \ref{alg:active_flow} (in red), random sampling from Algorithm \ref{alg:random_flow} (in blue), hybrid sampling described through Section \ref{appendix:hybrid_g4_exp} with $N_1 = 20$ (in pink) and $N_1= 40$ (in purple)}
\label{fig:r2_mean_g4_hybrid}
\end{figure}

\begin{figure}
\centering
\includegraphics[width=.5\textwidth]{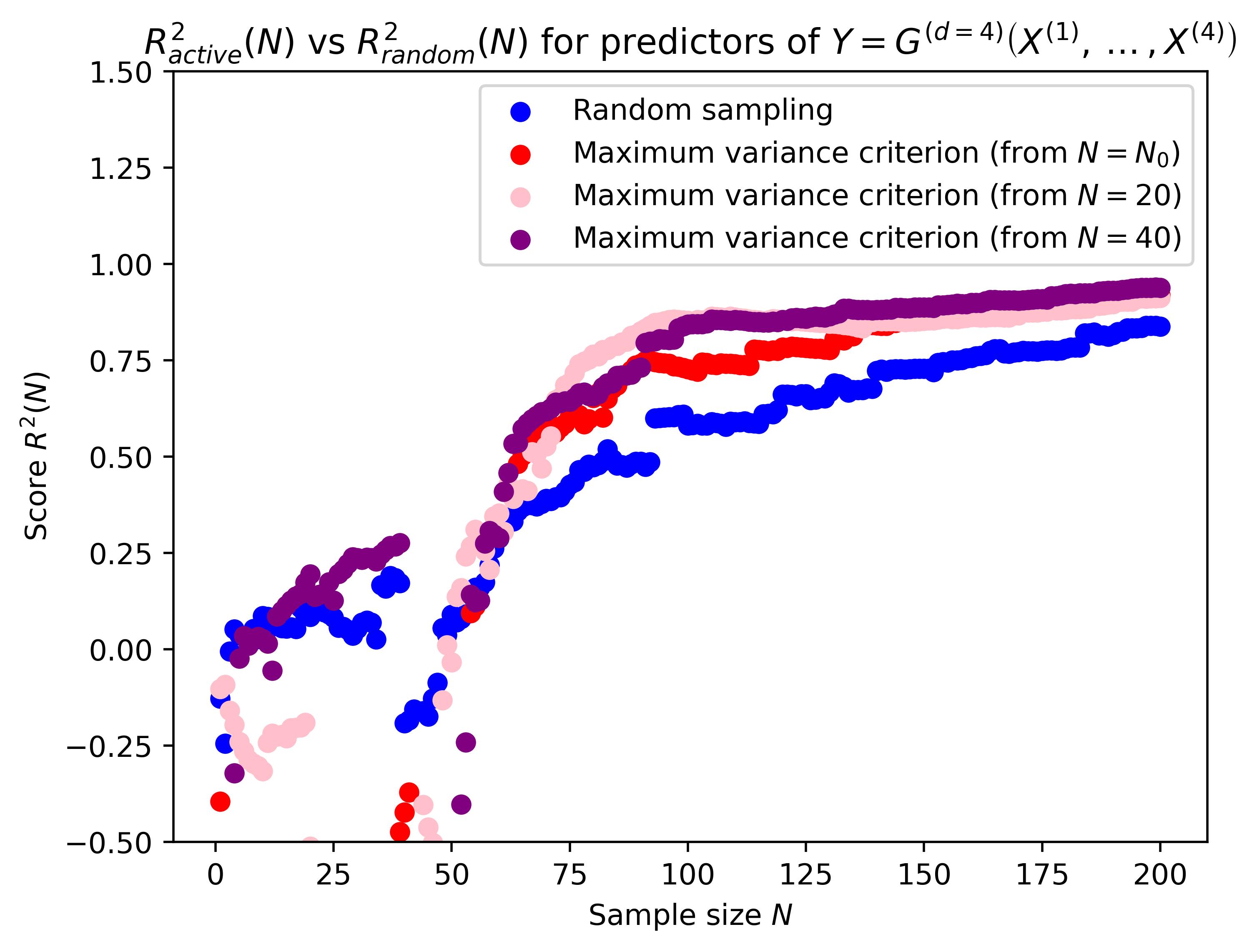}
\caption[$R^{2}(N)$ depending on some samplings methods for G-function (scaled)]{Same as Figure \ref{fig:r2_mean_g4_hybrid}, but with y-axis scaled between $-0.5$ and $1$.}
\label{fig:r2_mean_g4_hybrid_scaled}
\end{figure}

In order to minimize the phenomena described up there for small sample size, we try some hybrid sampling flow, that is~:
\begin{itemize}
    \item applying Algorithm \ref{alg:random_flow} from $S_{N_0}$ until $N_f = N_1$, to obtain $S_{N_1}^{\mathrm{random}}$;
    \item then, applying Algorithm \ref{alg:active_flow} from $S_{N_1}^{\mathrm{random}}$ until $N_f = N_2$, to obtain $S_{N_2}^{\mathrm{hybrid}}$.
\end{itemize}
As previously, we use the Gaussian Process Regressor generated to choose the next sample as model validation to compute our score, what can be observe on Figure \ref{fig:r2_mean_g4_hybrid} (scaled version available on Figure \ref{fig:r2_mean_g4_hybrid_scaled}). We can observe that we reduce the anomaly previously observed by our flow for small sample size. Waiting for another solution, it invites us to take an initial budget higher than $N_0 = 1$. \\

\subsubsection{Active Sampling VS Random Sampling for HSOTA}
\label{appendix:low_dim_hsota_exp}

We also study our score metric in another framework :
\begin{itemize}
\item $N_0 = 30$ and $N_f = 1 \ 000$
\item the expensive simulator is the the one simulating the slew up rate for HSOTA design;
\item for Algorithm \ref{alg:active_flow}, $m=2,5.10^4$ ;
\item the number of data in testing set is taken at $N_T = 5.10^3$.
\item as an oracle, we reduced the number of total features to the $10$ highlighted by FOCI with all the $30 \ 000$ samples, as mentioned in Section \ref{section:datasets}.
\end{itemize}

On Figure \ref{fig:r2_mean_hsota_low_dim}, we can observe the behavior of $R^2_{\mathrm{active}}(N)$, which seems outperform $R^2_{\mathrm{random}}(N)$.

\begin{figure}
\centering
\includegraphics[width=.5\textwidth]{figures/r2_mean_for_slewUp(hsota)_low_dim.jpg}
\caption[Evolution of $R^{2}_{\mathrm{active}}(N)$ vs $R^{2}_{\mathrm{random}}(N)$ for HSOTA]{We compute $R^{2}_{\mathrm{active}}(N)$ (in red) and $R^{2}_{\mathrm{random}}(N)$ (in blue) against $N$, as described through the framework of Section \ref{appendix:low_dim_hsota_exp}).}
\label{fig:r2_mean_hsota_low_dim}
\end{figure}